\documentclass[empty]{arxivtwocol}

\usepackage[T1]{fontenc}
\usepackage{times}%
\usepackage[square,sort,comma,numbers]{natbib}
\usepackage{amsmath}
\usepackage{amssymb}
\usepackage{bm}
\usepackage{dcolumn}
\usepackage{graphicx}
\usepackage{url}

\newcolumntype{d}[1]{D{.}{.}{#1}}

\firstpage{1} \lastpage{12} \volume{0} \pubyear{2022}

\begin{document}
\begin{frontmatter}                           

%
\title{DeepCreativity: Measuring Creativity with Deep Learning Techniques}

\runningtitle{DeepCreativity: Measuring Creativity with Deep Learning Techniques}

\author[A]{\fnms{Giorgio} \snm{Franceschelli}\thanks{Corresponding author. E-mail: giorgio.franceschelli@unibo.it.}} and
\author[B]{\fnms{Mirco} \snm{Musolesi}}
\runningauthor{G. Franceschelli and M. Musolesi}
\address[A]{Alma Mater Studiorum Università di Bologna, Italy\\
E-mail: giorgio.franceschelli@unibo.it}
\address[B]{University College London, United Kingdom,\\
	        The Alan Turing Institute, United Kingdom,\\
	        Alma Mater Studiorum Università di Bologna, Italy\\
E-mail: m.musolesi@ucl.ac.uk}

\begin{abstract}
Measuring machine creativity is one of the most fascinating challenges in Artificial Intelligence.
This paper explores the possibility of using generative learning techniques for automatic assessment of creativity. The proposed solution does not involve human judgement, it is modular and of general applicability. We introduce a new measure, namely \textit{DeepCreativity}, based on Margaret Boden's definition of creativity as composed by \textit{value}, \textit{novelty} and \textit{surprise}. We evaluate our methodology (and related measure) considering a case study, i.e., the generation of 19th century American poetry, showing its effectiveness and expressiveness.
\end{abstract}


\end{frontmatter}

\section{Introduction} \label{introduction}


Evaluation is a crucial concern in Artificial Intelligence and in science more in general. Measures and metrics are fundamental not only to check the validity of a hypothesis, but also to understand if it is possible to use some given results with confidence as a starting point for future research. An example is Shannon's entropy, which plays a central role as a measure of information, choice and uncertainty \cite{jha} and underpins many results in Information Theory~\cite{shannon}. In particular, the role of measures and metrics is crucial in machine learning, where the evaluation of algorithms for training and fine-tuning models is essential. An incredibly simple metric like accuracy, for example, is used in almost every machine learning project as a performance measure. The algorithms themselves are based on error measurements, such as the back-propagation algorithms \cite{rumelhart}, which relies on loss functions like mean squared error or cross-entropy.

It is not accidental that excellent progress has been made in benchmark tasks coupled with metrics used to quantify and to improve performance~\cite{rahwan}: examples include \cite{papineni} for machine translation or \cite{imagenet} for image generation. At the same time, there is also a need to derive new metrics for examining the behavior of algorithms in different environments and in relation with society \cite{rahwan}.
Among the spectrum of behaviors that could be exhibited by a machine, creativity is certainly one of the most interesting and one of the most important \cite{miller}. In fact, we have witnessed the emergence of an entire new field of research, namely Computational Creativity, with a focus on the study of the behaviors exhibited by artificial systems that would be deemed as creative \cite{wiggins2006, colton2012}. Indeed, one of the key goals of this field is the definition of evaluation techniques for measuring machine creativity.

In this paper, we present a novel methodology (and related measure) to evaluate the creativity of a generative agent. In particular, \textit{DeepCreativity} is based on the very famous definition of creativity provided by Margaret Boden: ``creativity is the ability to come up with ideas or artifacts that are new, surprising and valuable'' \cite{boden}. Although it is not the unique definition of creativity available (over one hundred of definitions have been proposed during time \cite{treffinger1996, aleinikov}), it is a fundamental one in the field of computational creativity, and there is a certain agreement of the importance of all of the aspects it takes into consideration. Our proposed measure uses deep learning techniques, in order to avoid the need of a human in the process, to measure how much an artifact is valuable, novel and surprising with respect to a given context, and therefore to measure the ability of an agent to come up with creative results. To the best of our knowledge, this is the first attempt to define an evaluation method for assessing creativity which is automatic and of general applicability.

Therefore, this work is structured as follows: a review of the literature about automatic methods to assess creativity is presented in Section \ref{relatedwork}; then, in Section \ref{creativitymeasure} the proposed creativity measure is discussed. An evaluation of \textit{DeepCreativity} is presented in Section \ref{experiments}, considering a case study of text generation in the context of 19th century American poetry; finally, we discuss limitations and potential future work in Section \ref{conclusions}. 

\section{Related Work} \label{relatedwork}

Over the years, several computational approaches have been proposed to automatically assess the creativity in products made by (human or artificial) agents, differing in the scope of evaluation or in the method. A complete survey can be found in \cite{franceschelli2021}. All of them consider value and novelty as aspects of creativity, while only some of them also consider surprise.
In the following, we will consider the three factors separately.

\subsection{Value} \label{valueworks}

Value, sometimes referred as quality, expresses how an artifact compares to others in its class in terms of utility, performance or attractiveness. It is typically defined as a weighted sum of performance attributes or as a reflection of the acceptance of the artifact by society \cite{maher2010}.
The authors of \cite{elgammal2015} follow the latter definition, which suggests to compute creativity by using an art graph where each vertex represents an artwork and each arc, connecting an older to a newer work, is labeled with the similarity between the two returned by an appropriate similarity function. The higher the similarity with subsequent works, the higher the value (and therefore, the higher the creativity). However, this method does not allow to compute value for the most recent works, but only for the older ones.
The former definition is more common in the literature. For instance, in \cite{maher2010} the authors suggest to derive value as the weighted sum of pre-defined performance variables. In \cite{maher2012}, value is defined using clusters of artifacts built on a performance space - with artifacts expressed as sets of attribute-value pairs. The authors of \cite{franca2016} define it as the synergy \cite{corning} between artifacts, expressed following the regent-dependent model. Also several domain-specific methods follow the definition of value as the sum of performance attributes or performance measures: for example, for poetry generation, the authors of \cite{yi2018} consider topic distribution (through LDA), fluency (through a neural language model) and coherence (through mutual information and TF-IDF) as components of value; in \cite{zhang2014} coherence is used (through BLEU, originally proposed for machine translation in \cite{papineni}) with quality (through perplexity); while the authors of \cite{yu2016} uses BLEU only.
However, the definition of value as the weighted sum of sub-components has the limitation of requiring the correct identification of all the relevant factors and their relative weights, which is a complex and time-consuming task.

\subsection{Novelty} \label{noveltyworks}

Novelty is commonly defined as the measure of how much an artifact differs from known artifacts in its class \cite{maher2010}. For this reason, a classic technique to measure novelty consists in the calculation of the distance between a given artifact and the other artifacts on a descriptive space, as discussed in \cite{maher2010} and \cite{maher2012}. The descriptive space is usually identified by the attributes used to define the artifacts. Similarly, domain-specific methods consider novelty in terms of distance or dissimilarity: for instance, in case of text generation, the authors of \cite{karampiperis} consider novelty as the average semantic distance between the dominant terms included in the textual representation of the story, compared to the average semantic distance of the dominant terms in all stories.  In~\cite{yi2018} diversity and innovation in poetry generation are measured by means of bigram-based average Jaccard similarity. As for value methods, the requirement of defining artifacts in terms of attributes appears to be a rather strong limitation.

A different definition of novelty has been proposed in \cite{berlyne}, namely as the degree an input differs from what an observer has experienced before. In \cite{elgammal2015} novelty is defined by considering the time dimension of personal experience: the lower the degree of similarity between an artifact and the previous works, the higher the novelty contribution of creativity. Even if not exactly used as an evaluation technique, a novelty score is proposed to guide the training of the generative part of the Creative Adversarial Network, a sort of \textit{creative-oriented} variant of the world-famous Generative Adversarial Network \cite{goodfellow2014}, is discussed in \cite{elgammal2017}. In addition to the classic adversarial loss provided by the discriminative model, the generator is trained to maximize a novelty loss that represents how much the generated artifact differs from previous works in terms of style. Although considering novelty as the deviation from style norms is somehow simplistic, it only requires a style classifier, automatically capturing an important aspect of novelty at the same time.

\subsection{Surprise} \label{surpriseworks}
In \cite{berlyne}, surprise is defined as the degree of disagreement between the real input and what it was expected in its place. This classic definition of surprise based on unexpectedness is typically also referred to as surprisal~\cite{tribus}. In \cite{maher2010}, unexpectedness is calculated considering whether or not the artifact follows the expected next artifact in the pattern recognized on recent artifacts. In \cite{grace2014}, surprise is measured as the unlikelihood of observing a particular artifact according to the predictions about relationships between its attributes. In the specific domain of text generation, in \cite{karampiperis}, surprise is defined as the average semantic distance between consecutive fragments of each story. For sequential artifacts like texts or sounds, the authors of \cite{bunescu} measure surprise considering the expected maximum surprise (as one minus the probability of the most unexpected token of the artifact) and the expected count of $\psi$-surprise (as the count of all the tokens which predictability is lower than a given threshold $\frac{\psi}{K}$), where the expectations are provided by an \textit{audience} neural network. In a similar way, \cite{macedo2004} proposes to quantify surprise considering both the probability of the event $X$ of interest and the probability of the most probable event $Y$, since the surprise of an event $X$ also depends on the certainty of $Y$ (e.g., ten equiprobable events have a very high unexpectedness, but should have a very low surprise, since we are not surprised to see one of them occurring).

A quite different approach is adopted in \cite{maher2012}, where the authors consider a new artifact as surprising if it creates a new cluster in the conceptual space (instead of perfectly fitting into an existing one). The idea of surprise as related with the difference between prior and posterior models is at the basis of Bayesian Surprise \cite{baldi2010}, used in \cite{franca2016} and \cite{varshney2013}. It is a measure of surprise in terms of the impact of a data point that changes a prior distribution into a posterior distribution, calculated applying Bayes' theorem (considering artifacts as a composition of attributes); here, surprise is the post-observation change rather than the prediction error.

\section{Measuring Creativity using Deep Learning} \label{creativitymeasure}

We now present \textit{DeepCreativity}, a new Deep Leaning creativity measure. The goal is to define a measure of more general applicability. Deep Learning is used for avoiding the need of identifying the required attributes to describe the artifacts or the components of creativity \cite{franceschelli2021}. This leads to a measure that allows for automatic evaluation of artifacts.
As discussed in Section \ref{introduction}, \textit{DeepCreativity} is based on the definition of creativity proposed by \cite{boden}. Therefore, the measure is based on three main factors, which will be explored in the next subsections separately: value (Subsection~\ref{value}), novelty (Subsection~\ref{novelty}) and surprise (Subsection~\ref{surprise}). Finally, in Subsection \ref{together}, we will put everything together by providing a unified definition of creativity.

\subsection{Value} \label{value}

We measure \textit{value} by means of the discriminative part of a Generative Adversarial Network \cite{goodfellow2014}. The GAN is trained by considering the real artifacts as the true ones; in this way, the discriminative model should learn a representation of real (and valuable) data, and its evaluation of a new artifact provides insights of its value in that context. Therefore, the value of an artifact $a$ over the value discriminator $D_v$ can be expressed as:

\begin{equation}
    V\!\left(a, D_v\right) = D_v\!\left(a\right),
\end{equation}

\noindent with $V\!\left(a, D_v\right)$ naturally constrained between $0$ (not valuable at all) and $1$ (highly valuable), since a sigmoid activation is applied to the output layer of $D_v$.

The choice of the real artifacts clearly influence the value measure proposed above. While it can be seen as a limitation of the approach, it is highly coherent with the nature of creativity itself. Creativity, and in particular value, are deeply \textit{context-dependent}: the same work, proposed in two different moments of history or to two different social groups may be evaluated differently \cite{boden}. Under this lens, the need of real artifacts conceals the opportunity of representing, within the measure, a fundamental aspect of creativity. The real data used during GAN's training will therefore represent a specific context, well-defined in temporal and cultural terms.

To train the GAN, it is important to distinguish between continuous tasks (like image generation) and sequential tasks (like text or sound generation).
With respect to continuous applications, a GAN can be trained using the following loss function \cite{goodfellow2014}:

\begin{equation}
\begin{split}
L &= \min_G \max_D V\!\left(D,G\right) = \mathbb{E}_{x \sim p_{data}\left(x\right)}[\log D\!\left(x\right)]\\[10pt]
 &+ \mathbb{E}_{z \sim p_z\left(z\right)}[\log \left(1 - D\!\left(G\!\left(z\right)\right)\right)],
\end{split}
\end{equation}

\noindent with $p_{data}$ as the real data distribution (representing the desired context), $p_z$ as the input noise variable, and with discriminator $D$ and generator $G$ trained alternately; notice that several refinements have been proposed in the recent years (see \cite{gui2020} for potential variations).

As far sequential applications are concerned, the impossibility of directly applying GAN to these tasks is a well-known problem \cite{goodfellow2016reddit, huszr2015}. A common way to solve it is by using SeqGAN \cite{yu2016}. SeqGAN considers the sequence generation process as a sequential decision making process, defining a reinforcement learning framework in which the generative model $G_{\bm{\theta}}$ is the agent, the actual state $\left(y_1, ..., y_{t-1}\right)$ is composed by the generated tokens so far, the next action $y_t$ is the next token to be generated, and the reward is the evaluation provided by the discriminative model $D_{\bm{\phi}}$. The generative model is then seen as a stochastic parametrized policy; Monte Carlo search is used to approximate the state-action value and directly train the policy via policy gradient \cite{yu2016}. More specifically, the REINFORCE algorithm \cite{williams1992} for learning the policy (but other methods can be used as well \cite{fedus2018}), which leads to the following update rule:

\begin{equation}
    \bm{\theta} \gets \bm{\theta} + \alpha Q_{D_{\bm{\phi}}}^{G_{\bm{\theta}}}\!\left(Y_{1:t-1}, y_t\right) \nabla_{\bm{\theta}} \ln G_{\bm{\theta}}\!\left(y_t | y_{1:t-1}\right),
\end{equation}

\noindent where $Q$ is the expected return obtained by the N-time Monte Carlo search.\par

\subsection{Novelty} \label{novelty}

With respect to \textit{novelty}, our definition is inspired by CAN \cite{elgammal2017} although the deviation from style norms cannot be used directly to measure the difference between artifacts. Therefore, as additionally done by the CAN discriminator, a neural network $D_n$ is trained to correctly recognize the style of real artifacts (from the given context). The neural network can just be a simple classifier (as in \cite{nam2019} for music or in \cite{tan2016} for paintings), outputting a probability vector of length $N$ equal to the number of possible classes. Consequently, a novelty measure can be defined as:

\begin{equation}\label{novmes}
\begin{split}
N\!\left(a, D_n\right) &= 1 - \frac{\sqrt{\sum_{i=1}^{N} \left( \frac{1}{N} - y_i \right) ^2}}{UB}\\[10pt]
 &\mbox{with} \quad UB = \frac{\sqrt{N \left( N-1 \right) }}{N},
\end{split}
\end{equation}

\noindent where $y$ is the output vector (of length $N$ and sum $1$) of $D_n$ given in input artifact $a$. The formula computes the Euclidean Distance between $y$ and the desired target vector of equiprobable values; in addition, it is constrained between 0 and 1, where it is equal to 1 when the distance is minimum (i.e., when the two vectors are equal) and it is equal to 0 when the distance is maximum (i.e., when a one-hot vector is considered). Please refer to Appendix \ref{proof} for the proof of this property.

\subsection{Surprise} \label{surprise}

With respect to \textit{surprise}, we follow the conceptual framework presented in \cite{baldi2010}. Starting from a sequential generative model $G_s$ trained to predict the next token given the previous ones on an appropriate training set (temporally and culturally defined, as stated for value), this allows for considering the impact of an artifact $a$ over $G_s$. Its influence is calculated using a weight correction applied over $G_s$ if $G_s$ is trained to correctly predict $a$. In analogy with the Bayesian Surprise, surprise is measured as the distance between prior $G_s$ (before training) and posterior $G_s$ (after training on $a$). The difference is in how the posterior distribution is obtained, namely not by means of Bayes' theorem, but through backpropagation and gradient descent. Notice that this idea is very close to the intrinsic reward presented in \cite{schmidhuber2010}, where a measure of surprise is derived by maximizing a distance function between prior and posterior distribution of a predictive model.

At inference time, only measuring surprise is relevant, and the model update is not actually required. It is only used to compute the weight correction $\Delta w_{ji}$, which expresses how much the posterior distribution will differ from the prior. Given an artifact $a = \{a_1, a_2, ..., a_N\}$, the mini-batch (of size $N$) gradient descent formula for $\Delta w_{ji}$ can be used:

\begin{equation}\label{deltaw}
    \Delta w_{ji} = - \eta \frac{1}{N} \sum_{k=1}^{N} \frac{\partial J_k}{\partial w_{ji}} ,
\end{equation}

\noindent where $\eta$ is the learning rate and $J_k$ is the loss function considering token $k$\footnote{It is worth noting that the loss function represents in a way the expectation error, i.e., the surprisal.}.

We can now define the surprise measure more formally. Given a sequential generative model $G_s$, an artifact $a$ has a surprise over $G_s$ equal to:

\begin{equation}
    S\!\left(a, G_s\right) = avg_{j, i}\left|\frac{\Delta w_{ji}}{w_{ji}}\right|.
\end{equation}

We note that the correction is divided by the weight to represent the degree of correction, i.e., the influence of the artifact. Then, the learning rate in Eq. (\ref{deltaw}) is not the learning rate used during training, but a parameter to adjust the magnitude of correction for the surprise measure. Even a value of 1 can be reasonable in certain problems. Finally, this approach requires $G_s$ in order to consider artifacts as sequential data, even if they are continuous. In case of image, $G_s$ may be, for instance, an autoregressive model (as in \cite{oord2016}, \cite{oord2016conditional} or \cite{parmar2018}).

\subsection{Putting All Together} \label{together}

Given the definition of $V\!\left(a, D_v\right)$, $N\!\left(a, D_n\right)$, $S\!\left(a, G_s\right)$ in the previous subsections, the \textit{DeepCreativity} measure (indicated with $DC$) is obtained by computing the creativity of a generative agent producing artifact $a$ over a temporal and cultural context $TCC$ as:

\begin{equation}
\begin{split}
    DC\!\left(a, TCC\right) = \; & \alpha_1 V\!\left(a, D_v\right)+\\[10pt]
    &\alpha_2 N\!\left(a, D_n\right) +\\[10pt]
    &\alpha_3 S\!\left(a, G_s\right),
\end{split}
\end{equation}
\noindent where $\alpha_1, \alpha_2, \alpha_3 \in \mathopen[0, 1\mathclose]$ and $\alpha_1 + \alpha_2 + \alpha_3 = 1$, and where $D_v$, $D_n$ and $G_s$ are trained over $TCC$, which is a set of examples $\left(x_1, y_1\right), ..., \left(x_n, y_n\right)$ where $x_1, ..., x_n$ are the real artifacts, and $y_1, ..., y_n$ are their labels representing the class (so we can assume $N$ different values). $\alpha_1, \alpha_2, \alpha_3$ weight the three single components of creativity; the immediate setting is to consider them as equal, as we will do in the following experiments. Nonetheless, it is possible to change them according to the specific domain, if some of the properties are found as more relevant in creativity assessment.

\section{Experiments} \label{experiments}

\begin{figure}[ht]
  \centering
  \includegraphics[width=1\linewidth]{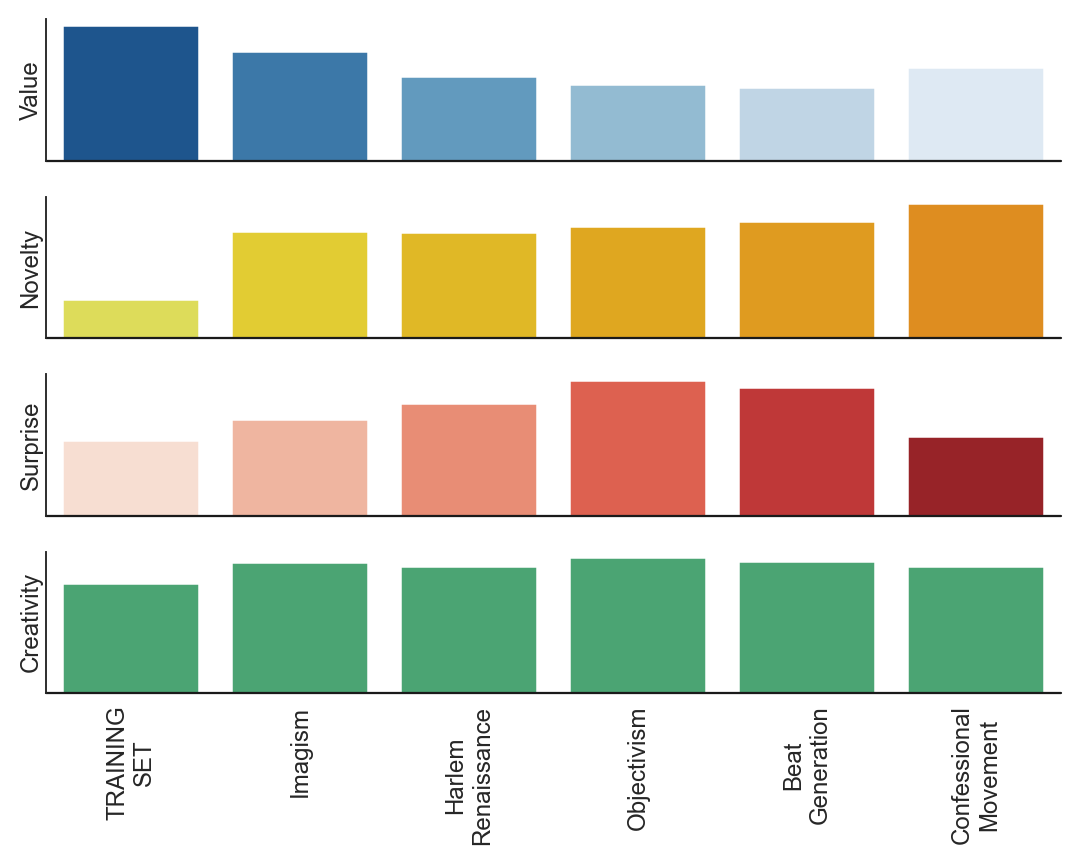}
  \caption{The average of value, novelty, surprise and creativity computed on a sample from the training set and on 20th century American poems.}
  \label{fig:vnsc}
\end{figure}

There is no common agreement about how to evaluate creativity measures. All the methodologies discussed in Section~\ref{relatedwork} have not been evaluated against a ground truth; on the contrary, they have just been tested over a generative system, in comparison with human judgements (always considering the products of a generative system) or they have not been tested at all. This can be attributed to the difficulty of finding a common definition of creativity, which is reflected in the lack of correct evaluation of creative productions.

However, a ground truth about this creativity process exist in this case: art history. The fact that in a certain moment of history, in a certain place, an artwork was appreciated or at least considered of sufficient quality to be ``printable'' may be used as useful information for evaluating a creative agent. Inspired by considerations done in \cite{miller} about CAN and its ability of intercepting the historical trajectory of art, a meta-evaluation test is defined, based on historical trajectories, to study if and how the proposed measure is able to correctly capture the changes of creativity over time in a fixed culture. In particular, the following experiment will concern the context of American Poetry.

The goal of this experiment is to measure the creativity of poems from different moment of history, but training the neural networks for the computation of \textit{DeepCreativity} on a specific historical context. \textit{DeepCreativity} can therefore be considered an appropriate creativity measure if the resulting creativity is higher for the artworks which really come after the context, because these are the ones been considered as highly creative in that moment. Consequently, it should also recognize the other works as less creative: later works should be judged more novel and surprising, but less valuable and understandable; and older works should be judged more admirable but less novel and surprising.
To verify this, two separated experiments are conducted, both of them with poems from 19th century as the context: the first one over poems from the 20th century, and the second one also considering poems from the 18th century (a sample of poems from the training set is always considered for a complete comparison). American poetry has been chosen because of the depth of available poems and the vastness of styles (i.e., poetic movements), but other contexts or arts could have been selected too. Extensive details about the data used are reported in Appendix \ref{dataset}; vice-versa, full details about implementation and training can be found in Appendix \ref{impldetails}.

With respect to the first experiment, Figure \ref{fig:vnsc} shows the average of the creativity components during movements and the final creativity measure. It is interesting to note that the higher the novelty the further from the training set. This correctly captures the fact that a movement, which immediately follows a certain period has to be novel with respect to it. Moreover, the next movement has to be novel with respect to both the works produced in that period and the first one. The surprise curve generally also shows a similar behavior: temporally distant artifacts are the result of different contexts and different situations and they are more difficult to be predicted only considering a \textit{past} version of the same culture. The last movement, the Confessional one, could be considered as an exception. This can be explained by considering how surprise is measured: in fact, it is calculated as the degree of change that the work causes over a 19th century American poems model, which is strictly related to a semantic view of the context, because it is based on the content. Indeed, temporally far movements might have a lower surprise measure if their themes (e.g., love) are semantically closer to those in the training set. 
The same consideration can be done to explain the value curve. For the first four movements, it tends to decrease with time, as expected. On the other side, Confessional Movement has a higher value; since its semantic content is closer to the one from the context, it results in a more similar and therefore comprehensible and admirable style, with a higher value.

In general, it is possible to observe that creativity tends to decrease further in time from the period of reference of the training set, while it is higher for the central movement, which is able to conciliate a high degree of surprise without a consistent loss in value.

\begin{figure}[ht]
  \centering
  \includegraphics[width=1\linewidth]{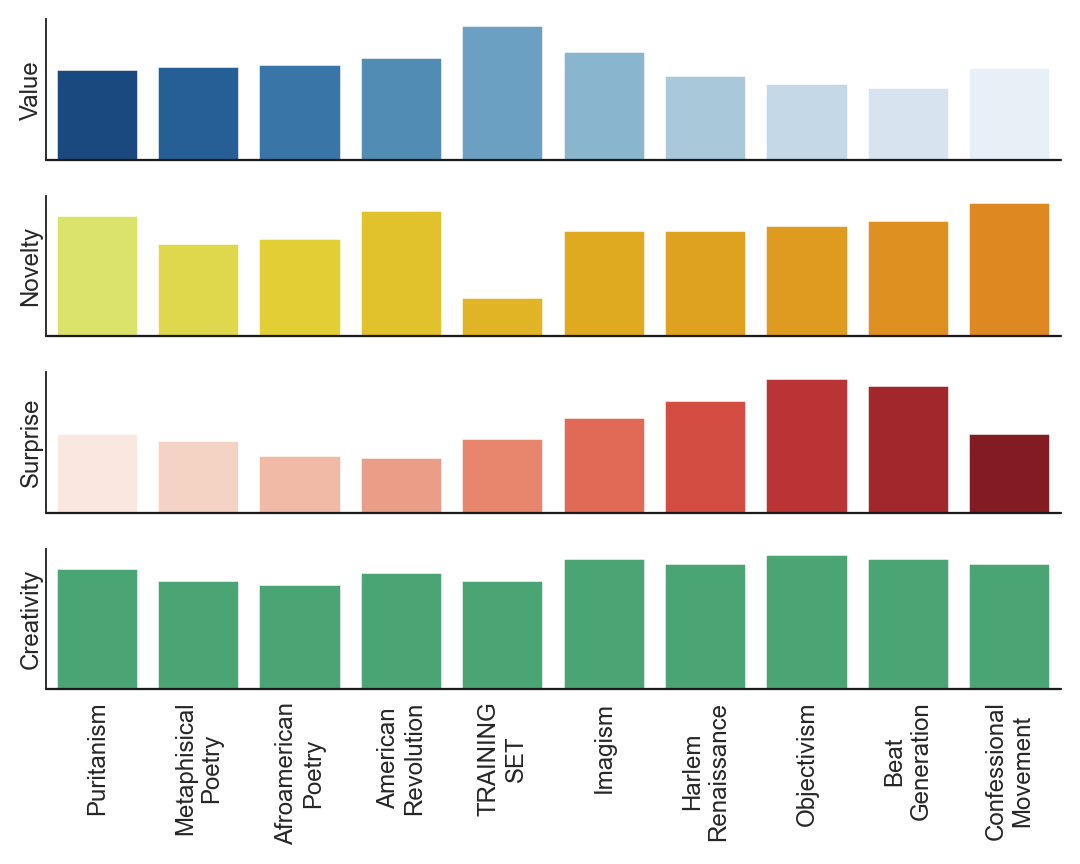}
  \caption{The average of value, novelty, surprise and creativity computed on a sample from the training set and on both 18th and 20th century American poems.}
  \label{fig:vnsc_prepost}
\end{figure}

With respect to the second experiment, Figure \ref{fig:vnsc_prepost} shows the three components also considering previous movements. This should help study the appropriateness of the three measures over the time dimension. It is therefore interesting to note that the curves follow the same trends observed for the subsequent century. Novelty and surprise generally increase further away from the period of reference of the training set; at the same time, their value decreases. In addition, it is interesting to note that surprise is smaller than for the 20th century on average. This is because the 19th century poems include some knowledge about the previous poems, making them more predictable.

\section{Conclusion and Future Work} \label{conclusions}

In this work, we have introduced \textit{DeepCreativity}, a new creativity measure based on three components with the objective of measuring the value, novelty and surprise of a generative process or algorithm, in terms of their products. This general approach overcomes the limits of having measures specific for certain domains; in addition, the use of deep learning techniques overcomes the limits of having to manually define the attributes or the components which characterize creativity. Finally, the need of a training set allows for the definition of a specific context of evaluation, which has been found to be a fundamental constraint of creativity. However, few limitations can also be found: novelty only considers the style or the genre, while it might lie in other traits of a work; and surprise requires a sequential generator, which could be not optimal for (supposedly simpler) continuous tasks.

The experiments conducted in the context of generative learning of 19th century American poetry have demonstrated that the measure is able to capture the historic trajectory of creativity over time, either only considering future poems or also previous ones, showing its effectiveness.
Additional tests should be carried out in order to confirm the correctness of the measure, ideally in different domains. 

\bibliographystyle{abbrv}
\bibliography{biblio}  

\cleardoublepage

\appendix

\section{Euclidean Distance Bounds with Proof} \label{proof}

The advantage of using Euclidean Distance is that it is bounded - both an upper and a lower bound can be derived. More formally:

\begin{equation}
    \sqrt{\sum_{i=1}^{N}\left(\frac{1}{N} - y_i\right)^2} \geq 0 ,
\end{equation}

where

\begin{equation} \label{lb}
\begin{split}
&\sqrt{\sum_{i=1}^{N}\left(\frac{1}{N} - y_i\right)^2} = 0\\[10pt]
 &\mbox{with} \quad y_i = \frac{1}{N}, i = 1,..,N,
\end{split}
\end{equation}

and

\begin{equation}\label{ub1}
    \sqrt{\sum_{i=1}^{N}\left(\frac{1}{N} - y_i\right)^2} \leq \frac{\sqrt{N\left(N-1\right)}}{N} ,
\end{equation}

where 

\begin{equation} \label{ub2}
\begin{split}
&\sqrt{\sum_{i=1}^{N}\left(\frac{1}{N} - y_i\right)^2} = \frac{\sqrt{N\left(N-1\right)}}{N}\\[10pt]
 &\mbox{with} \quad y_i = \begin{cases} 0, & \mbox{if } i = 1, ..., N, i \neq j \\ 1, & \mbox{if } i = j. \end{cases}
\end{split}
\end{equation}
\\
The derivation of Equation (\ref{lb}) is trivial (the lower bound of Euclidean Distance is $0$, and it is reached if and only if the two vectors are equal). In the following we present the derivation of Equations (\ref{ub1}) and (\ref{ub2}).

The squared sum of the difference can be decomposed as:

\begin{equation}
    \sum \left(x - y\right)^2 = \sum x^2 + \sum y^2 - 2 \sum x y .
\end{equation}
\\
Here, $x$ is a constant vector of $N$ values, each of them equal to $\frac{1}{N}$; for this reason,

\begin{equation}
\begin{split}
\sum x^2 &= \sum_{i=1}^N \left(\frac{1}{N}\right)^2 = \left(\frac{1}{N}\right)^2 \sum_{i=1}^N 1\\[10pt]
 &= \frac{1}{N^2} N = \frac{1}{N}.
\end{split}
\end{equation}
\\
Since $x$ is constant, it is possible to say that:

\begin{equation}
    2 \sum xy = 2 \sum_{i = 1}^{N} \frac{1}{N} y = \frac{2}{N} \sum_{i = 1}^{N} y ,
\end{equation}
\\
but $y$ is a vector of probabilities, therefore $\sum y = 1$, and so:

\begin{equation}
    2 \sum xy = \frac{2}{N} \sum_{i = 1}^{N} y = \frac{2}{N} .
\end{equation}
\\
Two of the three terms are constant; the last one depends on the variables of $y$. For a vector $y$ that has the property of $\sum y = 1$, the theoretical maximum of $\sum y^2$ is attained when all its entries are $0$ except one, which is 1, obtaining that:

\begin{equation}
    \sum y^2 = \sum_{i = 1}^{N} y_i^2 \leq 1 ,
\end{equation}
\\
where

\begin{equation}
\begin{split}
    &\sum_{i = 1}^{N} y_i^2 = 1\\[10pt]
    &\mbox{with} \quad y_i = \begin{cases} 0, & \mbox{if } i = 1, ..., N, i \neq j \\ 1, & \mbox{if } i = j. \end{cases}
\end{split}
\end{equation}
\\
The upper bound of the Euclidean Distance can be found rewriting:

\begin{equation} \label{eq1}
\begin{split}
&\sqrt{\sum \left(x - y\right)^2} = \sqrt{\sum x^2 + \sum y^2 - 2 \sum x y}\\[10pt]
&\quad \quad = \sqrt{\frac{1}{N} + 1 - \frac{2}{N}}= \sqrt{1 - \frac{1}{N}}\\[10pt]
 &\quad \quad = \sqrt{\frac{N - 1}{N}}=\frac{\sqrt{N\left(N-1\right)}}{N},\\[10pt]
 &\quad \quad \mbox{with} \quad y_i = \begin{cases} 0, & \mbox{if } i = 1, ..., N, i \neq j \\ 1, & \mbox{if } i = j. \end{cases}
\end{split}
\end{equation}

Equation (\ref{novmes}) can be finally obtained by setting $N = \frac{N - N_{min}}{N_{max}- N_{min}}$ and then $N = 1 - N$, which ensures that the measure is bounded between 0 and 1 and satisfies the desired properties listed in Subsection \ref{novelty}.

\newpage

\section{Details about Training Process and Implementation} \label{impldetails}

$G_v$ and $G_s$ are LSTM-based RNNs composed of an embedding layer of size 300 (with input length of 20), a LSTM layer with 256 units and a dropout of 0.2 rate, a dense layer with softmax activation~\cite{foster}. Following \cite{schmidt2020descending}, Adagrad \cite{duchi2011} with a learning rate equal to $0.01$ has been used as optimiser.
$D_v$ and $D_n$ are CNNs, implemented using an embedding layer of size 300 (with input length equal to the maximum poem length), three convolutional layers of 1 dimension (with tanh activation) with kernel sizes of, respectively, 3, 4 and 5, a max pooling over each output of the last convolutional layer, a dropout of 0.5 rate and finally a dense layer with softmax activation for $D_n$ and sigmoid activation for $D_v$~\cite{kim2014}. In this case, Adam \cite{kingma2014adam} with a learning rate equal to $0.0001$ has been adopted as optimizer. Word2Vec \cite{mikolov2013efficient} has been used for the embedding, pre-trained on Google News, and then fine-tuned for 100 epochs on the specific dataset.\par
$G_v$ and $D_v$ have been trained following the SeqGAN algorithm \cite{yu2016} except for the update rule followed, where REINFORCE with Baseline \cite{weaver2001} has been used in place of REINFORCE (with only positive rewards, it avoids to remain stuck in poor situations); $G_v$ has been pre-trained for 50 epochs on $TCC$, and $D_v$ for 5 epochs on batches of 32 outputs of $G_v$, with $N=1$ in the computation of the expected return of the Monte Carlo search, with $g_{steps} = 8$ and $d_{steps} = 4$, with a discriminator batch size of 32, and in total 550 epochs (with more, the discriminator becomes overfitted). 
$D_n$ has been trained for 56 epochs using Categorical Cross-Entropy; $G_s$ has been trained for 136 epochs using Sparse Categorical Cross-Entropy. In both cases, the number of epochs has been found as the one which minimizes the validation loss.\par
Finally, out-of-vocabulary (OOV) words of test set poems have been treated by substituting each OOV word with the one that has the most similar embedding to the one predicted by a sequence generator \cite{meemulla} - here, $G_s$. 

\newpage

\section{American Poetry Dataset} \label{dataset}

The training set is composed by 2676 poems from the 19th century, divided into five groups: American Renaissance (Brahmins and Romantics); Local Color; Naturalism; Neogothic (or Protodecadentism). For Brahmins, we included poems written by Henry Wadsworth Longfellow (extracted by \textit{The complete poetical works of Henry Wadsworth Longfellow}), Oliver Wendell Holmes (extracted by \textit{Songs in many keys}), James Russell Lowell (extracted by \textit{Poems}) and John Greenleaf Whittier (extracted by \textit{Poems of nature plus poems subjective and reminiscent and religious poems}, \textit{Anti-slavery poems and songs of labor and reform}, \textit{Personal poems}). For Romantics, we included poems written by Emily Dickinson (extracted by \textit{Poems}), Walt Whitman (extracted by \textit{Leaves of grass}) and Ralph Waldo Emerson (extracted by \textit{Poems}). For Local Color, we included poems written by Bret Harte (extracted by \textit{East and west: poems}), Frances Harper (extracted by \textit{Poems}) and Rose Terry Cooke (extracted by \textit{Poems}). For Naturalism,  we included poems written by Stephen Crane (extracted by \textit{The black riders and other lines} and \textit{War is kind}) and Hamlin Garland (extracted by \textit{Prairie songs}). For the last group, we included poems written by Edgar Allan Poe (extracted by \textit{The complete poetical works of Edgar Allan Poe}). Since all of these works are in public domain, we downloaded them from Project Gutenberg\footnote{\url{https://www.gutenberg.org/}} without any risk of copyright infringement.

The first test set, instead, is divided into five poetical movements of the 20th century. These movements are temporally consecutive, and they model a sort of timeline of American poetry. Of course, many other movements could be considered beside the five presented here, but they overlap the chosen five in time, making it difficult to interpret the results; in addition, no representativeness has been lost in our opinion, since the goal is not to retrace every single historically relevant moment in American poetry. Instead, the goal is to check if the defined measure is able to capture the concept of creativity for a certain period of time.
We took into consideration a timeline covering the following movements: Imagism; Harlem Renaissance; Objectivism; Beat Generation; and Confessional Movement. For each movement, 23 poems were considered. For Imagism, the poems are \textit{Autumn}, \textit{The embankment}, \textit{Above the dock}, \textit{Conversion} (by T.E. Hulme); \textit{Comraderie}, \textit{Piazza San Marco}, \textit{Ballad for gloom}, \textit{A song of the virgin mother}, \textit{Grace before song} (by Ezra Pound); \textit{Leda}, \textit{Pursuit}, \textit{Gift}, \textit{The shepherd}, \textit{All mountains} (by Hilda Doolittle); \textit{Sex and trust}, \textit{Furniture}, \textit{Censors}, \textit{After school}, \textit{Autumn sunshine} (by D.H. Lawrence); \textit{After all}, \textit{A night piece}, \textit{Modern love}, \textit{The feather} (by Ford Madox Ford). For Harlem Renaissance, the poems are \textit{Living earth}, \textit{Skyline}, \textit{Words for a hymn to the sun}, \textit{Banking coal}, \textit{Men}, \textit{Peers} (by Jean Toomer); \textit{God}, \textit{I look at the world}, \textit{Madam and the movies}, \textit{Silly animals}, \textit{Blues fantasy}, \textit{God to hungry child} (by Langston Hughes); \textit{The expulsion of hagar}, \textit{In memory of Arthur Clement Williams}, \textit{An offering}, \textit{Judith}, \textit{Ode to the sun}, \textit{Early spring} (by Eloise A. Bibb); \textit{Idolatry}, \textit{My heart has known its winter}, \textit{A note of humility}, \textit{Nocturne of the wharves}, \textit{Miracles} (by Arna Wendell Bontemps). 
For Objectivism, the poems are \textit{Brilliant sad sun}, \textit{The aftermath}, \textit{The dish of fruit}, \textit{Spring}, \textit{Flowers by the sea}, \textit{A goodnight} (by William Carlos Williams); \textit{A clerk tiptoeing the office floor}, \textit{Death of an insect}, \textit{Episode in Iceland}, \textit{The doctor's wife}, \textit{Lesson of job}, \textit{Hardly a breath of wind} (by Charles Reznikoff); \textit{Ode to the commonplace}, \textit{Aubade}, \textit{In the first circle of limbo}, \textit{Museum}, \textit{Eye to eye}, \textit{Ode on arrival} (by Carl Rakosi); \textit{I'm the worse for drinking again}, \textit{Poor soul! softly, whisperer}, \textit{Gin the goodwife stint}, \textit{Darling of gods and men}, \textit{Isn't it poetical a chap's mind?} (by Basil Bunting). 
For the Beat Generation movement, the poems are \textit{Homeless compleynt}, \textit{Ego confession}, \textit{G.S. reading poesy at Princeton} (by Allen Ginsberg); \textit{The leaves danced to Mozart}, \textit{Roma}, \textit{I saw great neptune}, \textit{So much depends upon}, \textit{I saw two lovers} (by Lawrence Ferlinghetti); \textit{Doctors will be protected}, \textit{Vows}, \textit{Out west}, \textit{Why California will never be like Tuscany}, \textit{Lodgepole} (by Gary Snyder); \textit{Storm at low tide}, \textit{The gash}, \textit{No sound}, \textit{Good morning}, \textit{The sacred distillate} (by William Everson); \textit{Getting to the poem}, \textit{Inter and outer rhyme}, \textit{Daydream}, \textit{Destiny}, \textit{Sunrise} (by Gregory Corso). 
For the Confessional Movement, the poems are \textit{Stopped dead}, \textit{The night dances}, \textit{A secret}, \textit{Cut}, \textit{Amnesiac}, \textit{You're}, \textit{The moon and the yew tree} (by Sylvia Plath); \textit{Buying the whore}, \textit{The red dance}, \textit{Old}, \textit{Not so, not so}, \textit{The fury of jewels and coal}, \textit{The fallen angels}, \textit{June bug} (by Anne Sexton); \textit{The moth chorale}, \textit{Parents}, \textit{Phone message}, \textit{Lasting}, \textit{Cherry saplings}, \textit{Leavings} (by W.D. Snodgrass); \textit{The cage}, \textit{Letter to his brother}, \textit{Rock-study with wanderer} (by John Berryman).
Apart from T.E. Hulme's poems, which are in the public domain, the others are retrieved through ProQuest's Literature Online database, by means of the license agreement of the University of Bologna. The poems were copied only for the amount of time required for their use in the experiments, and then deleted; in this way, their usage is in compliance with the current legislation, as explained in~\cite{margoni}.

The second test set is divided into four periods: Puritanism (Colonial age), in the 17th century; Metaphisical Poetry, at the beginning of the 18th century; the birth of African-American Poetry, at the half of the 18th century; and the American Revolution, at the end of the 18th century. For each period, one poet is considered: for the first one, Anne Bradstreet with the poems \textit{In memory of my dear grandchild Elizabeth Bradstreet}, \textit{In thankful remembrance}, \textit{For deliverance from a feaver}, \textit{Davids lamentation for Saul and Jonathan}, \textit{What God is like to him I serve}, \textit{In my solitary hours in my dear husband his absence}, \textit{An apology}, \textit{Upon some distemper of body}, \textit{Spirit}, \textit{The vanity of all wordly things}, \textit{Deliverance from another sore fit}, \textit{Meditations divine and moral}, \textit{Upon a fit of sickness}, \textit{My thankfull heart with glorying tongue}, \textit{As spring the winter doth succeed}, \textit{The author to her book}, \textit{As weary pilgrim, now at rest}, \textit{Another}, \textit{To my dear children}, \textit{The four elements}, \textit{Here follows some verses upon the burning of our house}, \textit{The flesh and the spirit}, \textit{We may live together}; for the second one, Edward Taylor, with the poems \textit{1}, \textit{6}, \textit{29}, \textit{32}, \textit{38} and \textit{39} from \textit{Preparatory Meditations - First Series}, \textit{7}, \textit{12}, \textit{62}, \textit{143} and \textit{146} from \textit{Preparatory Meditations - Second Series}, \textit{The experience}, \textit{The Souls Groan to Christ for Succour}, \textit{Upon wedlock, and death of children}, \textit{Head of a white woman winking}, \textit{The wrong way home}, \textit{Upon a wasp chilled with cold}, \textit{Ebb and flow}, \textit{The joy of church fellowship rightly attended}, \textit{Huswifery}, \textit{Upon a spider catching a fly}, \textit{The souls admiration hereupon}, \textit{The souls address to Christ against these assaults}; for the third one, Phillis Wheatley, with the poems \textit{On virtue}, \textit{To the University of Cambridge, in New-England}, \textit{On the death of a young lady of five years of age}, \textit{On the death of a young gentleman}, \textit{To a lady on the death of her husband}, \textit{Thoughts on the works of Providence}, \textit{To a lady on the death of three relations}, \textit{To a clergyman on the death of his lady}, \textit{An hymn to the morning}, \textit{An hymn to the evening}, \textit{On recollection}, \textit{On imagination}, \textit{To a lady on her coming to North-America with her son, for the recovery of her health}, \textit{To a lady on her remarkable preservation in an hurricane in North-Carolina}, \textit{To a lady and her children, on the death of her son and their brother}, \textit{On the death of J. C. an infant}, \textit{To S. M. a young African painter, on seeing his works}, \textit{A farewell to America}, \textit{On the death of Dr. Samuel Marshall}, \textit{To a gentleman and lady on the death of the lady's brother and sister, and a child of the name of Avis, aged one year}, \textit{A funeral poem on the death of C. E. an infant of twelve months}, \textit{On the death of the Rev. Dr. Sewell}, \textit{On being brought from Africa to America}; for the fourth one, Philip Freneau, with the poems \textit{A New-York tory}, \textit{To lord Cornwallis}, \textit{The vanity of existence}, \textit{To the memory of the brave americans}, \textit{The royal adventurer}, \textit{A speech - that should have been spoken by the King of the Island of Britain to his Parliament}, \textit{Lines - occasioned by Mr. Rivington's new titular types to his Royal Gazette}, \textit{A prophecy}, \textit{The argonaut - or, lost adventurer}, \textit{Barney's invitation}, \textit{Sir Guy Carleton's address to the americans}, \textit{Scandinavian war song}, \textit{The projectors}, \textit{A picture of the times}, \textit{Satan's remonstrance}, \textit{The refugee's petition to Sir Guy Carleton}, \textit{To a concealed royalist}, \textit{The prophecy of king Tammany}, \textit{Stanzas - occasioned by the departure of the British from Charleston}, \textit{On the british king's speech}, \textit{Manhattan city}, \textit{A news-man's address}, \textit{The happy prospect}. Since all of these works are in public domain without risks of copyright infringement, they were downloaded from Project Gutenberg.

\end{document}